\documentclass[letterpaper, 10pt, conference]{ieeeconf}

\IEEEoverridecommandlockouts
\usepackage{graphicx}
\usepackage{times,cite,flushend}

\usepackage{amsmath,amssymb}
\usepackage{booktabs,array,multirow,makecell}
\usepackage{placeins}
\usepackage{microtype}
\usepackage{xcolor}
\usepackage[draft]{hyperref}

\newcommand{\todoev}[1]{}

\makeatletter
\let\exobridge@ieeemakecaption\@makecaption
\long\def\@makecaption#1#2{%
  \ifx\@captype\@IEEEtablestring
    \footnotesize\bgroup\par\centering
    {\normalfont\footnotesize #1:\nobreakspace #2}\par\egroup
    \vspace{2pt}%
  \else
    \exobridge@ieeemakecaption{#1}{#2}%
  \fi}
\makeatother

\title{\LARGE\bf
ExoBridge: Learning a Bare Hand to Hand-Worn\\
Exoskeleton Mapping through Human Limb Coupling}

\author{Ruitong Tian$^{1}$, Xianyao Li$^{1}$, Noah B. Wilson$^{2}$, Fang Xu$^{1}$, and Eric Jing Du$^{1,\dagger}$%
\thanks{$^{1}$Ruitong Tian, Xianyao Li, Fang Xu, and Eric Jing Du are with the Department of Civil and Coastal Engineering, University of Florida, Gainesville, FL, USA.}%
\thanks{$^{2}$Noah B. Wilson is with the Department of Mechanical and Aerospace Engineering, University of Florida, Gainesville, FL 32611, USA.}%
\thanks{$^{\dagger}$Corresponding author: Eric Jing Du (\texttt{eric.du@essie.ufl.edu}). First author: Ruitong Tian (\texttt{ruitongtian@ufl.edu}).}%
}

\begin{document}
\raggedbottom
\maketitle
\thispagestyle{empty}
\pagestyle{empty}
\bstctlcite{exobridge:BSTcontrol}

\begin{abstract}
Human video offers a scalable source of experience for dexterous robot learning, but obtaining motion and tactile supervision while preserving bare hand interaction remains challenging. We present ExoBridge, a framework that leverages human limb coupling to learn a bridging function from bare hand video to the motion and tactile state of a sensorized exoskeleton. Our central idea is to use coordinated bimanual behavior to connect an uninstrumented visual source with a measured manipulation interface. During collection, one hand remains bare and provides visual observations, while the opposite hand wears the exoskeleton and supplies synchronized motion and tactile measurements. These paired demonstrations train a temporal visual model to predict fingertip contact, continuous tactile intensity, and relative encoder motion from bare hand video alone. The exoskeleton defines an intermediate state space whose motion coordinates are linked to a dexterous robot hand through existing calibration. Evaluation on 1,215 demonstrations across four manipulation tasks uses held out collection sessions and yields a pooled any contact AUROC of 0.916 and a Pearson correlation of 0.790 for tactile intensity. The learned bridge also predicts relative changes in exoskeleton configuration from bare hand video. These results demonstrate that human limb coupling can turn exoskeleton measurements into supervision for bare hand video, establishing a learned bridge between human visual demonstrations and a robot oriented manipulation interface.
\end{abstract}

\section{Introduction}

Human manipulation is a useful source of experience for dexterous robot learning, and video can capture it without requiring a robot during every demonstration. Recent methods use human observations together with motion supervision to learn representations and policies that transfer across embodiments \cite{wang2024dexcap,kareer2024egomimic,zheng2026egoscale,hoque2026egodex,yuan2025motiontrans}. Bare hand video is particularly attractive because the observed hand can interact with objects without wearing a sensing interface \cite{shi2025zeromimic}. However, the resulting images do not directly provide the encoder state and tactile measurements recorded by an interface designed for a robot hand. Connecting these visual observations to a measured manipulation state remains an important step between collecting human video and using it for robot learning.

\begin{figure*}[t]
\centering
\includegraphics[width=\textwidth]{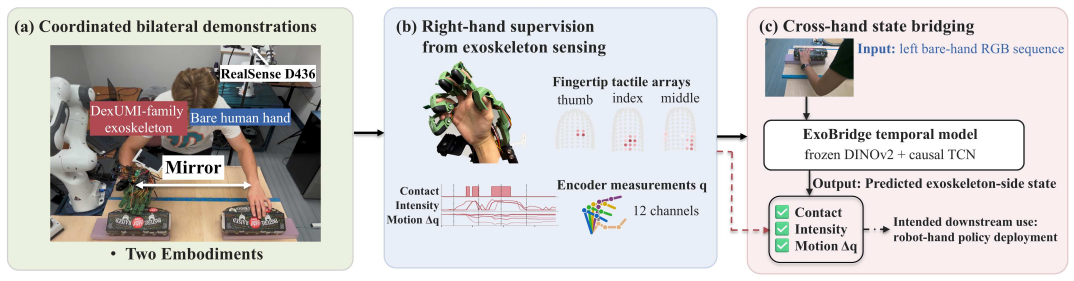}
\caption{Overview of ExoBridge. (a) Coordinated bilateral demonstrations: the bare left hand is observed while the sensorized right hand performs the same task on a separate object copy. (b) Exoskeleton sensing on the right hand supplies the fingertip tactile arrays and encoder measurements that serve as training targets. (c) ExoBridge maps the bare hand RGB sequence to the predicted exoskeleton state: contact, tactile intensity, and relative motion in the exoskeleton state space.}
\label{fig:overview}
\end{figure*}

Motion and tactile supervision address different parts of this problem. Hand tracking and retargeting connect visible motion to a robot configuration, while visual tactile learning associates images with measured contact or pressure \cite{wang2024dexcap,grady2023pressurevisionpp,zhao2024egopressure,handa2020dexpilot,qin2022onehand,pavlakos2024hamer}. A sensorized exoskeleton brings both signals into one collection interface. DexUMI records encoder motion and fingertip tactile measurements during direct interaction, with an existing mapping between exoskeleton configuration and a target robot hand \cite{xu2025dexumi}. The exoskeleton therefore provides a useful target space. The question is how to learn this state from a visual stream in which the observed hand remains bare.

Our central idea is to use human limb coupling to construct supervision between the two hands. Studies of intermanual coordination describe dependencies between the timing and organization of coordinated actions \cite{swinnen2002intermanual,gerloff2002bimanual,kelso1984phase,cardoso2002neuronal}. We use this principle to pair two roles during data collection. One hand supplies the visual observation, while the opposite hand supplies motion and tactile measurements. Rather than prescribing an identical joint trajectory for both hands, we learn the predictive relationship produced by their coordinated execution of the same manipulation objective \cite{mechsner2001perceptual}. This leads to the central question of this work: \textbf{How can human limb coupling be used to learn a bridging function from bare hand video to exoskeleton motion and tactile state?}

We introduce ExoBridge, shown in Fig.~\ref{fig:overview}, to learn this mapping from coordinated bimanual demonstrations. The bare left hand interacts with one object copy within the camera view. The right hand wears a sensorized exoskeleton from the DexUMI family and performs a coordinated instance of the same task on a separate copy \cite{xu2025dexumi}. Synchronized recordings pair the bare hand video with the exoskeleton measurements. A temporal visual model learns a tactile mapping that predicts fingertip contact and continuous tactile intensity, together with a motion mapping that predicts changes in encoder configuration relative to the first sampled state. At inference, the model uses only the bare hand RGB stream. The measured target space gives the bridge a concrete output definition, while the existing configuration calibration connects its motion component to a chosen robot hand.

We evaluate the proposed bridge across egg carton opening, lid twisting, soldering tool orientation, and USB insertion. Recordings from each collection session remain together during training and evaluation, so the reported predictions are obtained on held out sessions. Contact discrimination, continuous intensity prediction, and relative motion estimation test complementary outputs of the same mapping framework. The results show predictive contact and intensity information across all four tasks, together with relative motion correspondence that is strongest in active channels. An aligned sequence then illustrates how these outputs describe the same manipulation episode.

Our contributions are threefold.
\begin{itemize}
\item A cross-hand supervision framework that leverages human limb coupling to learn a mapping from bare hand RGB video to the tactile and relative motion state of a contralateral sensorized exoskeleton.
\item An empirical characterization of predictively transferable interaction information across coordinated hands, demonstrating contact discrimination, continuous tactile intensity prediction, and relative encoder motion prediction on held out sessions across four manipulation tasks.
\item A paired dataset of 1,215 coordinated bimanual demonstrations with synchronized RGB, tactile, and encoder measurements, together with session-level splits and evaluation code for reproducible study of cross-hand state prediction.
\end{itemize}

\section{Related Work}

\subsection{Human Demonstrations for Dexterous Robot Learning}

Human demonstration interfaces provide motion information that can connect human experience to robot control. UMI records manipulation through a portable gripper interface, while DexCap captures wrist and finger motion alongside visual observations and retargets the motion to a robot hand \cite{chi2024umi,wang2024dexcap}. EgoMimic learns from egocentric human and robot demonstrations, and EgoScale uses motion supervision and aligned human and robot data to support transfer from human video \cite{kareer2024egomimic,zheng2026egoscale}. These approaches make the representation connecting human observations and robot actions an explicit part of the learning problem \cite{xie2025human2robot}. ExoBridge studies a related mapping at the state estimation level. Its output includes measured fingertip interaction as well as relative motion, and the supervision is supplied by an exoskeleton worn on the opposite hand.

\subsection{Wearable Interfaces and Exoskeleton State}

Wearable interfaces allow an operator to demonstrate manipulation through direct interaction while recording motion for a target robot \cite{yang2024ace,fangDEXOPDeviceRobotic2025}. DexUMI uses a hand exoskeleton to reduce kinematic mismatch and record encoder and tactile signals \cite{xu2025dexumi}. ExoStart uses sensorized exoskeleton demonstrations to initialize robot learning, while DexEXO studies wearability and hardware alignment for demonstration collection \cite{si2025exostart,zhu2026dexexo,du2025mile,ohkawa2026yubi,chen2026dexvitac}. These systems place the wearable interface in the demonstration process to obtain measurements related to robot manipulation. In ExoBridge, the exoskeleton instead provides the supervised target for a separate bare hand visual stream. Its encoder coordinates define the motion output, and its fingertip sensors define the tactile output. This separates the hand that is observed from the hand that provides the measurements while retaining a target space associated with a robot interface.

\subsection{Visual Tactile Estimation and Bilateral Supervision}

Visual tactile estimation connects hand appearance and motion to measured interaction \cite{grady2022pressurevision,jeon2026hope}. PressureVision++ estimates fingertip pressure from RGB images, and EgoPressure studies pressure and hand pose from egocentric observations \cite{grady2023pressurevisionpp,zhao2024egopressure}. EgoTac predicts tactile signals from egocentric video, while TouchAnything studies bimanual tactile estimation using visual observations and synchronized sensing \cite{zhang2026egotac,zhou2026touchanything}. These methods motivate learning contact and continuous sensor responses from visual input, with the target representation determined by the available measurements \cite{narasimhaswamy2020contacthands,yang2022touchandgo}.

Obtaining supervision while retaining bare hand appearance is a separate collection problem \cite{yin2025osmo}. Glove2Hand synthesizes bare hand imagery from sensing glove recordings while retaining the associated physical measurements \cite{zhang2026glove2hand}. EgoTactile provides a bilateral alternative by pairing a visible bare hand with a synchronized gloved hand outside the camera view \cite{zeng2026egotactile}. ExoBridge uses coordinated bimanual supervision to learn a different target: the relative encoder motion and fingertip tactile state of a sensorized exoskeleton during manipulation. The proposed bridging function is therefore defined by both the supervision relationship and the exoskeleton state space to which the video is mapped.

\section{ExoBridge}

\subsection{Problem Formulation}

Each demonstration pairs an RGB sequence of the bare hand with synchronized measurements from the contralateral exoskeleton. Let \(\mathbf{X}_{1:t}=\{x_1,\ldots,x_t\}\) denote the visual history through sample \(t\). We learn the bridging function
\[
B_\theta[\mathbf{X}_{1:t}]
=
\left[
\hat{c}^{\mathrm{any}}_t,
\hat{\mathbf{c}}_t,
\hat{\mathbf{s}}_t,
\widehat{\Delta\mathbf{q}}_t
\right],
\]
where \(\hat{c}^{\mathrm{any}}_t\) is the probability of contact on any instrumented fingertip, \(\hat{\mathbf{c}}_t\) contains the three per fingertip contact probabilities, \(\hat{\mathbf{s}}_t\) contains three continuous tactile intensities, and \(\widehat{\Delta\mathbf{q}}_t\) is the relative encoder motion. Each output refers to the exoskeleton state at the corresponding sample. Coordinated bimanual demonstrations supply the paired observations and targets used to learn this function.

The tactile mapping associates visible hand object interaction with the measured contact and aggregate fingertip response. The motion mapping associates the visible action with changes in exoskeleton configuration. These targets share a synchronized recording, but describe different aspects of the interaction. Contact identifies whether an instrumented fingertip is engaged, intensity represents the magnitude of the defined sensor response, and relative encoder motion represents configuration change. Their definitions remain in the exoskeleton measurement space throughout training and evaluation.

\subsection{Coordinated Bimanual Demonstrations}
\label{sec:collection}

\begin{figure}[t]
\centering
\includegraphics[width=\columnwidth]{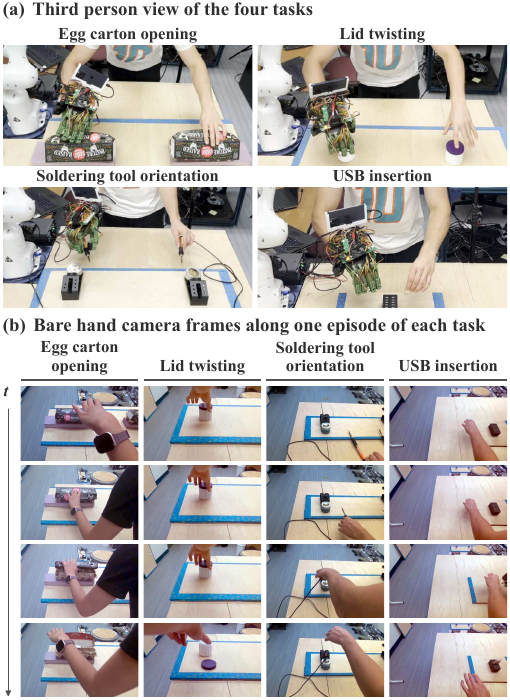}
\caption{Coordinated bimanual demonstrations. (a) Third person view of the four tasks. (b) Bare hand camera frames along one episode of each task, top to bottom in time order.}
\label{fig:collection}
\end{figure}

A fixed RealSense D436 camera positioned near the operator viewpoint records the bare left hand manipulating one copy of the task object (Fig.~\ref{fig:collection}). The right hand wears the sensorized exoskeleton and performs a coordinated instance of the same manipulation objective on a separate copy. The hands coordinate task execution without a prescribed identical joint trajectory. The exoskeleton remains outside the visual model view. This arrangement preserves bare hand appearance in the input while obtaining measured supervision during the same demonstration. In Fig.~\ref{fig:overview}(a), Mirror marks the mirrored placement of the two object copies in front of the operator, and Two Embodiments refers to the bare hand and the exoskeleton hand.

The exoskeleton records twelve encoder channels and tactile signals from the thumb, index finger, and middle finger \cite{xu2025dexumi}. Each of these fingertips contains 120 taxels. RGB, encoder, and tactile streams are aligned on the recorded timeline. The model uses every fourth aligned sample, giving an effective sampling rate of approximately 5 Hz. The collection covers four forms of manipulation: enclosure opening and release in egg carton opening, sustained rotational contact in lid twisting, pickup and reorientation in soldering tool orientation, and precision grasping and alignment in USB insertion. These tasks provide different interaction settings for evaluating the same learned mapping.

Table~\ref{tab:dataset} summarizes the final usable dataset. It contains 1,215 episodes from 31 distinct collection sessions. Aligned step counts in Table~\ref{tab:dataset} are reported before temporal subsampling. A session can include multiple tasks, so task session counts overlap while the total counts distinct sessions. The any contact column is the proportion of samples with a positive contact target on at least one instrumented fingertip. Session identifiers are retained across tasks and used to separate training and evaluation recordings. Demonstrations were recorded from experts (IRB number anonymized for review). A session is a consecutive block of demonstrations recorded in one sitting.

\begin{table}[t]
\caption{Paired bimanual recordings by task.}
\label{tab:dataset}
\centering
\footnotesize
\setlength{\tabcolsep}{3.5pt}
\begin{tabular}{@{}lrrrr@{}}
\toprule
Task & Episodes & \makecell[r]{Task\\sessions} & \makecell[r]{Aligned\\steps} & \makecell[r]{Any\\contact} \\
\midrule
Egg carton opening & 245 & 18 & 44,404 & 26.6\% \\
Lid twisting & 340 & 20 & 56,157 & 35.0\% \\
Soldering tool orientation & 311 & 20 & 57,221 & 38.7\% \\
USB insertion & 319 & 20 & 46,804 & 35.8\% \\
\midrule
\textbf{Total} & \textbf{1,215} & \textbf{31} & \textbf{204,586} & \textbf{34.4\%} \\
\bottomrule
\end{tabular}
\end{table}

\subsection{Exoskeleton State Targets}
\label{sec:targets}

For fingertip \(f\) and taxel \(m\), let \(r_{t,f,m}\) denote the recorded normal tactile signal. We reference the signal to the first sampled frame \(t_0\) of each episode,
\[
z_{t,f,m}=r_{t,f,m}-r_{t_0,f,m}.
\]
The per fingertip contact target applies a fixed threshold to the maximum referenced taxel response,
\[
c_{t,f}
=
\mathbf{1}\left[\max_m z_{t,f,m}>\tau\right],
\qquad \tau=10~\mathrm{LSB}.
\]
The any contact target is the union of the three fingertip targets,
\[
c_t^{\mathrm{any}}
=\max_{f\in\mathcal{F}}c_{t,f},
\qquad
\mathcal{F}=\{\mathrm{thumb},\mathrm{index},\mathrm{middle}\}.
\]
Continuous tactile intensity retains the aggregate response of each fingertip in logarithmic form,
\[
s_{t,f}
=
\log\left[1+\max\left[0,\sum_m z_{t,f,m}\right]\right].
\]
The sum is taken over the referenced taxels and then truncated at zero. This definition produces one scalar intensity target per instrumented fingertip. The full taxel measurements remain available in the recording, while the learned tactile output consists of contact probabilities and these three scalar intensities.

We define motion relative to the initial sampled configuration,
\[
\Delta\mathbf{q}_t=\mathbf{q}_t-\mathbf{q}_{t_0}.
\]
This representation expresses the configuration change within an episode in the fixed reference encoder coordinates of the collection system. The coordinates are not anatomical joint angles. The model predicts all twelve encoder channels. The primary motion evaluation uses channels 0 through 9. Channels 10 and 11 are excluded from the primary motion summary because their robot mapping differs: one command follows the ring finger, while the distal command is fixed rather than independently calibrated. Channel 10, the active little finger channel, is reported in Section~\ref{sec:motion}.

\subsection{Temporal Visual Model}
\label{sec:model}

\begin{figure*}[t]
\centering
\includegraphics[width=\textwidth]{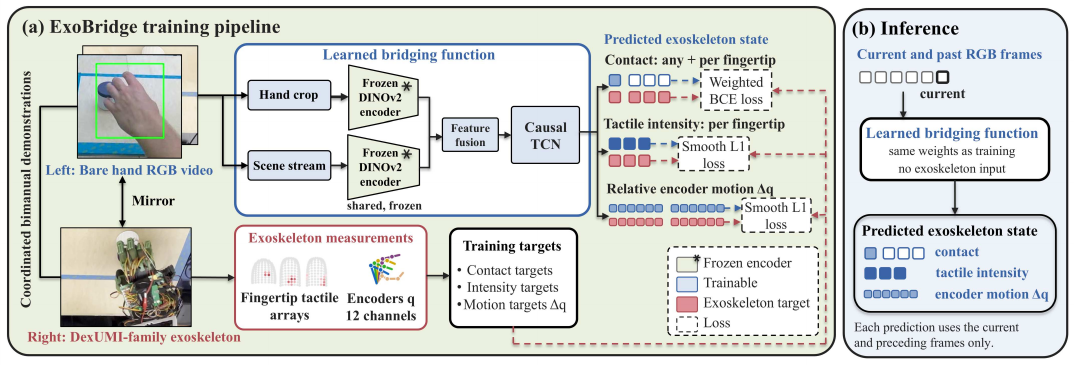}
\caption{ExoBridge training pipeline and inference. (a) The bare hand video and the paired exoskeleton measurements train the bridging function; each output is supervised by its exoskeleton target. (b) Inference uses the current and past RGB frames only.}
\label{fig:pipeline}
\end{figure*}

The visual model represents local hand interaction and scene context using two streams. MediaPipe Hands supplies a hand bounding box, which is expanded by a 25 percent margin to form the hand crop \cite{zhang2020mediapipehands}. The second stream masks the detected hand region and retains the surrounding scene. Frozen DINOv2 ViT-S/14 features encode both views \cite{oquab2023dinov2}. Normalized bounding box coordinates are added to the hand representation. The two streams are projected to a shared 512-dimensional space and combined before temporal processing. This provides local appearance and spatial context as inputs to the same prediction problem. Fig.~\ref{fig:pipeline} summarizes the training pipeline: the bare hand video is split into the hand crop and the scene stream, both are encoded by the frozen DINOv2 features and fused, and the causal temporal model produces the predicted exoskeleton state. Each loss compares a prediction with the corresponding exoskeleton target, and inference uses the same model with only the current and past RGB frames.

Six residual temporal convolution blocks process the fused sequence \cite{bai2018tcn}. They use kernel size three, dilations of 1, 2, 4, 8, 16, and 32, and causal padding, and the causal receptive field spans 127 sampled frames. Prediction heads estimate any contact, per fingertip contact, continuous tactile intensity, and relative encoder motion. The contact heads use class weighted binary cross entropy. Intensity and motion use Smooth L1 regression after standardization with statistics from the training partition. The model is trained on the four tasks jointly. Training uses AdamW for 30 epochs with a learning rate of \(3\times10^{-4}\), weight decay 0.01, and batches of at most 16 complete episode sequences, and the reported predictions come from the final epoch without validation based checkpoint selection. Exoskeleton measurements supply the training targets, while inference uses the bare hand RGB stream without exoskeleton input.

\subsection{Evaluation Protocol}
\label{sec:eval}

We use five folds separated by collection session. All recordings from one session remain in the same fold across tasks, and each held out episode is predicted by a model trained without that session. The procedure is repeated with three random seeds. Task results evaluate the jointly trained model on each task, while pooled results combine the held out tasks within each fold. Each metric is computed on the pooled samples of a held out fold, averaged over the five folds, and then averaged over the three seeds, as reported in Table~\ref{tab:transfer}. The episode figures show the seed 0 prediction. This protocol evaluates prediction on new collection sessions within the recorded task set.

Any contact is evaluated using the area under the receiver operating characteristic curve, abbreviated as AUROC. Continuous tactile intensity is evaluated using Pearson correlation in the logarithmic target space, normalized mean absolute error, and mean absolute error after inversion to recorded sensor units. Normalization uses the range between the fifth and ninety fifth percentiles of the held out targets. The sensor error is reported in LSB, the digital sensor unit. Correlation measures agreement in the continuous target, while the error measures quantify differences in predicted magnitude.

Motion is evaluated with mean absolute error in relative encoder coordinates and Pearson correlation for each channel. A zero change reference predicts \(\Delta\mathbf{q}_t=\mathbf{0}\) throughout the episode. Its error measures the magnitude of motion away from the initial state and provides a direct comparison with remaining at that state. We report the reference beside the learned prediction so that motion error can be interpreted relative to the change present in each task.

\section{Experiments}

We evaluate whether the proposed bridging function predicts measured exoskeleton state from bare hand video in held out collection sessions. The contact experiment examines the supervision relationship and discrimination of the measured contact target. Continuous tactile intensity evaluates prediction of the aggregate fingertip response. Relative motion compares the predicted configuration change with a static initial state reference. Finally, an aligned demonstration illustrates the tactile and motion outputs together. Table~\ref{tab:transfer} summarizes the quantitative results under the common model configuration, folds, and seeds.

\begin{table*}[t]
\caption{Held out prediction of exoskeleton tactile state and relative motion (encoder channels 0 to 9), pooled over five session folds and three seeds.}
\label{tab:transfer}
\centering
\footnotesize
\setlength{\tabcolsep}{5pt}
\begin{tabular}{@{}lcccccc@{}}
\toprule
Task & \makecell{Any contact\\AUROC $\uparrow$} & \makecell{Intensity \(r\)\\$\uparrow$} & \makecell{Intensity\\nMAE $\downarrow$} & \makecell{Intensity MAE,\\LSB $\downarrow$} & \makecell{Motion\\MAE $\downarrow$} & \makecell{Zero change\\MAE $\downarrow$} \\
\midrule
Egg carton opening & 0.915 & 0.724 & 0.166 & 12.7 & 3.36 & 5.02 \\
Lid twisting & 0.888 & 0.737 & 0.189 & 15.1 & 4.58 & 7.38 \\
Soldering tool orientation & 0.922 & 0.821 & 0.140 & 19.7 & 4.78 & 9.70 \\
USB insertion & 0.939 & 0.856 & 0.116 & 10.6 & 3.53 & 7.33 \\
\midrule
\textbf{All tasks} & \textbf{0.916} & \textbf{0.790} & \textbf{0.153} & \textbf{14.8} & \textbf{4.15} & \textbf{7.51} \\
\bottomrule
\end{tabular}
\end{table*}

\subsection{Contact Supervision and Prediction}

The contact evaluation first asks whether the collection protocol pairs measured exoskeleton contact with visible interaction on the bare hand. Of 196 held out frames with completed blind labels, 85 were tactile positive under the exoskeleton threshold, and 81 of these, 95.3 percent, showed visible bare hand contact, with a 95 percent frame level bootstrap percentile interval from 90.2 to 98.9 percent over 2,000 replicates \cite{efron1979bootstrap}. The labels were assigned from the bare hand frames alone, without access to the exoskeleton measurements. This is a conditional correspondence check: among samples labeled positive by the exoskeleton, most also show contact in the visual stream. It supplies evidence about the supervision relationship independently of model prediction.

The learned mapping then tests whether this relationship supports contact discrimination in held out recordings. ExoBridge achieves a pooled any contact AUROC of 0.916, with task values from 0.888 to 0.939. Thus, the bare hand visual stream distinguishes the defined exoskeleton contact and noncontact targets across the four manipulation settings. The annotation and prediction results answer complementary questions. The annotation checks whether positive measurements are associated with visible interaction, while the model evaluation establishes how well the learned bridge discriminates the measured target in sessions excluded from training.

The recorded bilateral correspondence carries information beyond the task template. To rule out that the model exploits only task identity and coarse task progress rather than the recorded pairing of the two hands, we train a decoupled pairing condition in which each exoskeleton target sequence is replaced by another episode of the same task aligned by normalized progress. Discrimination is scored as pair AUC, the probability that a contact observation receives a higher predicted contact probability than a noncontact observation from the same task and the same normalized progress bin. True pairing reaches a pair AUC of 0.84 on egg carton opening, 0.82 on lid twisting, 0.75 on soldering tool orientation, and 0.80 on USB insertion, while decoupled pairing reaches 0.66, 0.58, 0.67, and 0.68, a difference of 0.18, 0.24, 0.08, and 0.12 whose paired session bootstrap intervals exclude zero for every task and seed. A predictor that uses only normalized task progress ranges from 0.48 to 0.60 in the same comparison. These comparisons use the contact output of the model reported in Table~\ref{tab:transfer}, averaged over its three seeds.

\subsection{Continuous Tactile Intensity Prediction}

The continuous target evaluates whether the bridge predicts the aggregate fingertip response rather than producing a contact probability alone. Across all tasks, intensity prediction reaches a Pearson correlation of 0.790, a normalized mean absolute error of 0.153, and an absolute error of 14.8 LSB. These results characterize agreement with the continuous exoskeleton measurement over the evaluated samples. They establish the intensity output of the bridge in the same sensor space used to define its supervision.

Correlation and magnitude error provide different information about this output. Soldering tool orientation has a higher intensity correlation than lid twisting, 0.821 compared with 0.737, but also a larger sensor unit error, 19.7 compared with 15.1 LSB. The normalized error in Table~\ref{tab:transfer} provides a comparison relative to the target range for each task. A stronger association with the target does not by itself imply a smaller absolute sensor error. Reporting these measures together describes both the correspondence learned by the bridge and the magnitude error that remains.

\begin{figure}[t]
\centering
\includegraphics[width=\columnwidth]{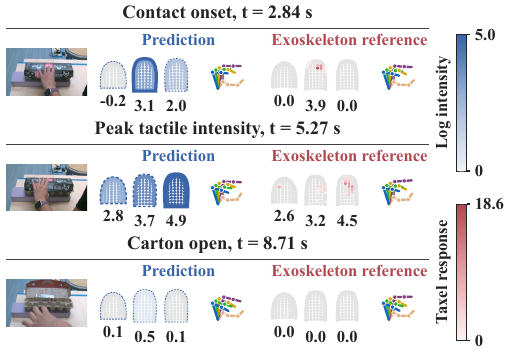}
\caption{Predicted and measured exoskeleton state at three instants of one held out egg carton demonstration: bare hand frame, prediction, and exoskeleton reference. Predicted pads visualize scalar fingertip intensity uniformly and do not represent taxel level prediction; the prediction and reference pads use separate color scales.}
\label{fig:states}
\end{figure}

\subsection{Relative Encoder Motion Prediction}
\label{sec:motion}

The motion experiment asks whether bare hand video predicts changes in the coordinated exoskeleton configuration more accurately than retaining its initial state. Across channels 0 through 9, the pooled mean absolute error is 4.15, compared with 7.51 for the zero change reference. This corresponds to a 44.7 percent reduction in error. The learned prediction also improves over the reference in each task, as shown in Table~\ref{tab:transfer}. The comparison establishes that the motion mapping estimates configuration change rather than relying on a static initial configuration.

\begin{figure}[t]
\centering
\includegraphics[width=\columnwidth]{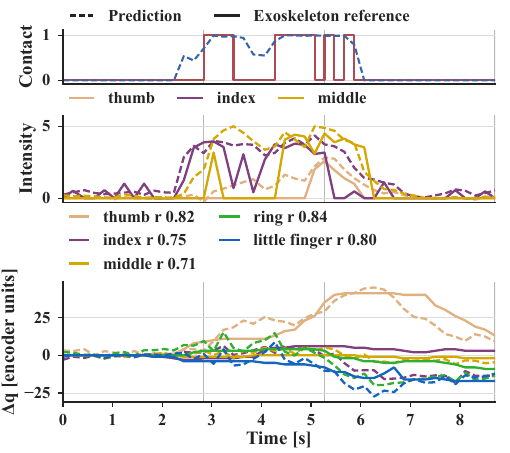}
\caption{Temporal comparison for the same held out episode: any contact, tactile intensity, and relative motion of five encoder channels. Dashed lines are predictions, solid lines the exoskeleton reference; legend correlations are from the full held out evaluation.}
\label{fig:traces}
\end{figure}

Correspondence is not uniform across encoder channels. Pooled over the held out sessions, the thumb, index, middle, ring, and little finger channels, encoder channels 1, 4, 6, 8, and 10, reach correlations of 0.82, 0.75, 0.71, 0.84, and 0.80. The twelve channels comprise three thumb channels, three index channels, and two channels for each of the middle, ring, and little fingers. For each finger we report the channel with the largest zero change reference error, that is, the channel whose configuration changes most within an episode: channels 1, 4, 6, 8, and 10, with zero change errors of 11.2, 17.2, 10.0, 26.0, and 25.6 against 0.9 to 3.4 for the other channels of the same fingers. The pooled error of Table~\ref{tab:transfer} covers channels 0 through 9, so channel 10 is reported here and excluded from that average. The reference error provides the scale of that motion, while the model error and correlation describe how well it is predicted. For the held out episode of Fig.~\ref{fig:states}, Fig.~\ref{fig:traces} shows these channels with the pooled correlation of each channel in its legend. The strongest motion evidence therefore concerns the active configuration changes of the coordinated manipulation.

\subsection{Tactile and Motion Outputs in a Held Out Demonstration}

Figs.~\ref{fig:states} and~\ref{fig:traces} illustrate how the learned bridge represents one held out egg carton episode. The example aligns bare hand video with contact, continuous tactile intensity, and relative motion on the same recording timeline. Three instants identify contact onset, peak measured intensity, and the final sampled state after the carton is open and the hand has withdrawn. The aligned trajectories provide the temporal context for these instants and show the outputs of the two mappings during the same interaction.

For motion visualization, the predicted relative encoder state is added to the measured initial exoskeleton configuration and passed through the existing exoskeleton to XHand calibration and forward kinematics. The measured configuration is rendered using the same calibration. Both renderings therefore share the measured initial state, and their difference reflects the predicted configuration change. In Fig.~\ref{fig:states}, the pads show the thumb, index, and middle fingertips from left to right. Each predicted pad carries a uniform shading given by the predicted scalar fingertip intensity and an outline weight given by the predicted contact probability, while each measured pad shows the individual taxel responses. The two sides use different color scales; the numbers under the pads are the predicted and measured scalar intensity. Fig.~\ref{fig:traces} adds vertical rules at the three instants of Fig.~\ref{fig:states}. Contact probabilities are aligned with the measured contact targets. This example explains the behavior of the mapping within an episode, while Table~\ref{tab:transfer} provides the evaluation across held out recordings.

\section{Discussion}

ExoBridge uses the two hands for different roles within one supervised learning process. The observed hand preserves the bare hand visual input, while the sensorized hand supplies the target measurements. Human limb coupling motivates the coordinated collection protocol, and the bridging function learns the resulting relationship between these streams. This provides a way to train a visual mapping into an exoskeleton measurement space without placing the exoskeleton on the hand seen by the model.

\begin{figure}[!t]
\centering
\includegraphics[width=\columnwidth]{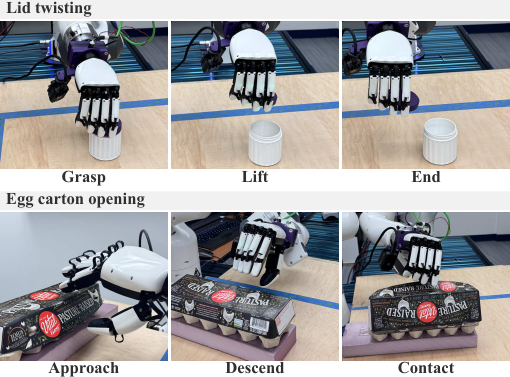}
\caption{Autonomous rollouts of policies trained on exoskeleton demonstrations: lid twisting (top) and egg carton opening (bottom). Frames run left to right in capture order and may come from different rollouts.}
\label{fig:rollouts}
\end{figure}

The target space gives the mapping both tactile and motion components. Contact and intensity describe measured fingertip interaction, while relative encoder motion describes configuration change. The existing calibration provides a connection from exoskeleton configuration to a robot hand, making the learned motion output relevant to subsequent robot transfer. Policies can be learned in this state space. In a companion experiment of this work, we train policies with a spatial tactile input \cite{huang20243dvitac} on demonstrations from the same exoskeleton. These policies reach success rates of 0.80 on egg carton opening and 0.55 on lid twisting, each over 20 autonomous rollouts with a 12-DoF XHand on a Franka Emika arm (Fig.~\ref{fig:rollouts}). They consume exoskeleton demonstrations directly rather than bridge predictions, so they establish the usefulness of the target space rather than closed loop use of the predicted state. The present evaluation establishes the prediction stage. A downstream policy study can then test the value of using the predicted state for robot learning and control.

We will release paired RGB recordings, tactile and encoder measurements, collection session identifiers, split assignments, and evaluation code. Video distribution will follow the applicable participant consent and data sharing requirements. These materials will support reproduction of the session separated evaluation and comparison of alternative implementations of the bridging function.

\section{Limitations}

The evaluation covers four tabletop tasks and one fixed camera arrangement. The reported generalization is to held out collection sessions within this setting. Tactile targets are measurements from the contralateral exoskeleton, and motion is evaluated in its recorded encoder coordinates. The all sample intensity evaluation does not isolate amplitude variation during contact. The experiments evaluate state prediction offline; the reported rollouts use exoskeleton demonstrations rather than bridge predictions, so closed loop use of the predicted state remains untested. Broader collection conditions and targeted control evaluations would extend the scope of the mapping established here.

\section{Conclusion}

We introduced ExoBridge, a framework that leverages human limb coupling to learn a bridging function from bare hand video to sensorized exoskeleton state. Coordinated bimanual demonstrations provide supervision for fingertip contact, continuous tactile intensity, and relative encoder motion while leaving the observed hand uninstrumented. Evaluation across four manipulation tasks demonstrates prediction of these measured targets in held out collection sessions, with the strongest motion correspondence in active channels. ExoBridge establishes a learned connection from bare hand visual observations to an exoskeleton state space containing both motion and tactile interaction measurements.

\section*{Acknowledgment}

The authors used AI assistants including Anthropic Claude Code, OpenAI ChatGPT, and OpenAI Codex for drafting and editing manuscript text, citation formatting, and the generation of figure and analysis code for Sections III to V. All text, analysis code, numerical results, and interpretations were reviewed and verified by the authors.

\FloatBarrier
\bibliographystyle{IEEEtran}
\bibliography{refs}

\end{document}